\documentclass[11pt]{article}
\usepackage[preprint]{acl}
\usepackage{times}
\usepackage{latexsym}
\usepackage[T1]{fontenc}
\usepackage[utf8]{inputenc}
\usepackage{microtype}
\usepackage{inconsolata}
\usepackage{graphicx}
\usepackage{booktabs}
\usepackage{amsmath,amssymb}
\usepackage{tabularx}
\usepackage{array}
\usepackage{enumitem}
\usepackage{dblfloatfix}
\usepackage{placeins}
\usepackage{float}
\newcolumntype{Y}{>{\centering\arraybackslash}X}
\newcolumntype{Z}{>{\raggedright\arraybackslash}X}

\title{Artifact-to-Capability: Diagnosing Breaks in Capability Formation of Self-Evolving Agents in a Controlled Environment}
\author{\normalfont Ziqian Ban$^{1}$, Haoxiang Yang$^{1}$, Yanfei Zhang$^{2}$ \\ Shixin Liang$^{1}$, Haoran Yu$^{1}$, Xu Lin$^{3}$ \\ Shaobin Zhao$^{1}$, Xuran Li$^{1}$, Zhiyi Cheng$^{1}$ \\[2pt] \normalfont $^{1}$Tianjin University \quad $^{2}$Independent Researcher \\ \normalfont $^{3}$International Digital Economy Academy (IDEA)}

\begin{document}
\maketitle

% Terminology (defined once, then used by name):
% artifact (persistent product of evolution); component (part of an artifact: skill, auto_player, code branch)
% frozen system (all artifacts that take part in decisions in a frozen version); mechanism (Reflexion / Voyager / DGM); version (DGM lineage node)
% state; move (action in the framework definitions); immediate winning point / immediate-win state / conflict state
% pass; pass rate; P(pass), P(pass | legal) = legal pass rate; newly passing / lost / net change; wins
% signal-based evaluation (three proxy signals); three-scale diagnosis = state-level / game-level / lineage-level diagnosis
% checkpoint (Properties 3+4); capability formation; capability chain; break
% capability-formation breaks: legalization gain (vs. selection gain), compositional override (via defense-first early return), loss-only drift
% Voyager configurations: full / -open3 / -minimax / round1

\begin{abstract}
Self-evolving agents write interaction experience into artifacts and reuse them in later tasks. Most work supports such methods with three proxy signals: artifacts accumulate, behavior changes once they are loaded, and metrics rise. We call this signal-based evaluation. What these signals are meant to indicate, however, has not been defined. To fill this gap, we define capability formation from artifacts as five sequentially dependent properties, two of which form a checkpoint for capability formation, and introduce three-scale diagnosis, which judges these properties at the state, game, and lineage levels. In freestyle Gomoku, Reflexion, Voyager, and DGM evolve with the same base model and all show the three signals, yet each exhibits a distinct capability-formation break. Reflexion shows a legalization gain: its pass rate rises only because more of its moves are legal. Voyager shows a compositional override: components that are sound on their own combine to miss all 20 available one-move wins. DGM shows loss-only drift: after passing the checkpoint, later versions only lose passing states. These counterexamples show that signal-based evaluation cannot determine whether a capability has formed.

\end{abstract}

\section{Introduction}

Self-evolving agents generate artifacts through interaction with an environment---textual reflections, executable skills, or code---and inject them into a harness so that experience can be reused across tasks~\citep{xi2023agents}. Work in this area typically argues that a method is effective using three signals: artifacts accumulate, behavior changes once they are loaded, and accuracy, pass rate, or evolutionary score rises. We call this \emph{signal-based evaluation}. All three signals are proxies for capability. None directly tests whether a changed behavior advances the task, whether an established effect is overturned by another part of the system, or whether later evolution builds on what has been gained. More fundamentally, what ``effective'' should refer to has not been defined. We therefore ask: what verifiable task capability do the artifacts a system generates and reuses actually leave behind?

\begin{figure*}[!t]
    \centering
    \includegraphics[width=\linewidth]{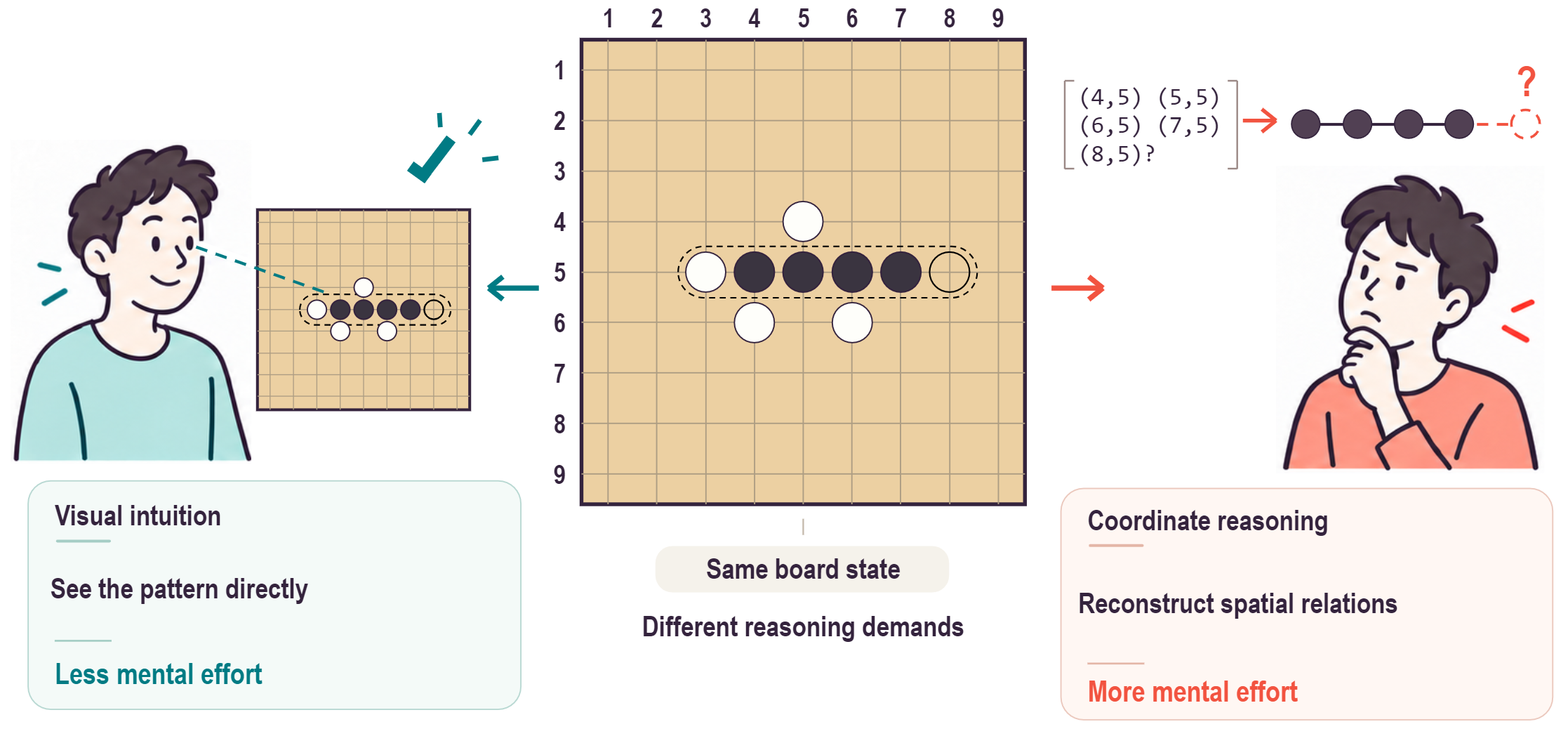}
    \caption{Different reasoning requirements from the same board state.}
    \label{fig:gomoku_spatial_reasoning}
\end{figure*}

Prior work has argued that endpoint performance on a single episode cannot characterize self-evolution, and has turned to long-horizon task streams, capability trajectories across frozen versions, or separate analyses of reflections, tool libraries, and transfer (Section~\ref{sec:related}). These analyses still take episodes, versions, or tools as their objects. Answering our question instead requires taking the transition from artifact to capability itself as the object: observing how a single action changes, whether that change survives in actual decision making, and whether it carries over across versions. Such observations are hard to make in open-ended tasks, where states cannot be replayed exactly, the correctness of an action lacks an independent criterion, and environmental noise and unstructured states confound attribution. We therefore study the question in a controlled environment.

Such an environment must make states replayable, task effects rule-judged, and games and versions fully recorded (Section~\ref{sec:env}). Inspired by~\citep{li2023emergent}, who study world-model representations with controlled board-game sequences, we use freestyle Gomoku (Gomoku without forbidden-move restrictions) as a ``microscope'' for self-evolving mechanisms. Gomoku meets these requirements. Simple rules do not make the task simple. The agent reads the board only as a sequence of coordinates and must reconstruct spatial relations along rows, columns, and diagonals on its own (Figure~\ref{fig:gomoku_spatial_reasoning}); without memory, about half of the model's moves land on occupied points, which leaves real room for evolution to improve.

In this environment, we define capability formation as five properties that an artifact passes through in sequence, forming the Artifact-to-Capability diagnostic framework (Section~\ref{sec:five-properties}). An artifact must first form and enter decision making, and then change actions. A change is neutral in itself: it must also produce a positive effect without undermining effects already established, and only when both hold has a capability formed. Later evolution should then extend it. The five properties are judged at three scales---states, games, and lineages---which we call \emph{three-scale diagnosis} (Section~\ref{sec:design}).

We let Reflexion~\citep{shinn2023reflexion}, Voyager~\citep{wang2023voyager}, and DGM~\citep{zhang2026dgm} evolve autonomously with the same base model, environment, and opponents, then freeze and diagnose them. To show that signal-based evaluation cannot establish capability formation, a single counterexample suffices: a system that shows all three signals but has not formed a capability. We present three, each breaking at a different property---three \emph{capability-formation breaks}. Reflexion shows a \emph{legalization gain}: its memories make the model play fewer illegal moves but do not make it choose better among legal ones. Voyager shows a \emph{compositional override}: each component is sound on its own, but in combination the calling script blocks the opponent and returns early at conflict states, overriding an established win; across 20 games, it completes five at none of its immediate-win states. DGM shows \emph{loss-only drift}: OB-3 passes the checkpoint, but later versions only lose passing states, and the last barely changes behavior. That the three breaks fall on different properties also shows that no property can stand in for another.

Our contributions are:
\begin{enumerate}
\item \textbf{Problem and definition.} We pose a question that signal-based evaluation leaves open---what verifiable task capability do the artifacts a system generates and reuses actually leave behind?---and define capability formation as five sequentially dependent properties, with task-relation realization and boundary-safe reuse forming the checkpoint for capability formation.
\item \textbf{Method.} We introduce three-scale diagnosis, which judges these properties at the state, game, and lineage levels in controlled freestyle Gomoku.
\item \textbf{Findings.} We present three counterexamples in Reflexion, Voyager, and DGM: all three signals are present, yet the diagnosis detects a legalization gain, a compositional override, and loss-only drift, breaking at task-relation realization, boundary-safe reuse, and capability extension, respectively. Signal-based evaluation therefore cannot determine whether a capability has formed.
\end{enumerate}

\section{Related Work}\label{sec:related}

Prior work has examined self-evolving agents and the roles of their artifacts from several angles. Process-oriented evaluations track performance changes along checkpoints or task streams. EvoPathBench freezes successive artifact checkpoints and uses paired evaluations with artifacts enabled and disabled on held-out episodes to test whether performance depends on them~\citep{lin2026evopath}. Agent Plasticity measures self-improvement at each checkpoint in environments including board games; using LLM-generated observational annotations, it distinguishes two bottlenecks---failure to reuse a relevant artifact and failure even after reuse---and notes that controlled interventions such as removal or replacement could further separate these mechanisms~\citep{singh2026plasticity}. \citet{lin2026harnessbenefit} trace the limited gains of weaker models to artifacts that are either not loaded or not followed after loading.

Other work documents negative effects of artifacts: when skills with overlapping functions coexist, conflicts arise mainly when the system selects which skill to read~\citep{yan2026oneskill}; as a skill pool keeps growing, its pass rate first rises and then falls~\citep{shang2026backfires}; and reflective memory can lower success relative to a memory-free retry in some settings~\citep{bai2026vrlbench}. These studies characterize change mainly through aggregate performance over tasks, episodes, or checkpoints; we refer to this as outcome analysis. Outcome analysis can reveal gains, regressions, and retention, and combined with interventions it can also test whether performance depends on artifacts.

Closer to the state level are studies that make paired comparisons on the same input. Misevolution compares action safety with and without memory for the same query~\citep{shao2026misevolve}; SkillApt compares answer correctness and cost with and without a skill in the same task state~\citep{guo2026skillapt}; Counterfactual Memory Audit crosses two histories while holding a verified-identical input fixed, distinguishing whether memory changes an action from whether it leads to the correct task branch~\citep{zhang2026cma}; and \citet{huang2024selfcorrect} classify transitions before and after self-correction on the same problem, including incorrect-to-correct and correct-to-incorrect changes.

Our object of diagnosis is capability formation itself: whether a frozen system can turn behavioral change into task capability. We decompose this transition into sequentially dependent properties (Section~\ref{sec:five-properties}) and pair each with independently assessable evidence: move differences and rule-based judgments in state-level diagnosis, call traces in game-level diagnosis, and newly passing and lost states in lineage-level diagnosis. Within this framework, the two bottlenecks identified by Agent Plasticity correspond to behavioral activation and task-relation realization, whereas the breaks in Reflexion, Voyager, and DGM fall at task-relation realization, boundary-safe reuse, and capability extension. Additional related work appears in Appendix~\ref{app:related}.

\section{Artifact-to-Capability Diagnostic Framework}
\subsection{Five Diagnostic Properties}\label{sec:five-properties}

\begin{figure*}[!t]
    \centering
    \includegraphics[width=\linewidth]{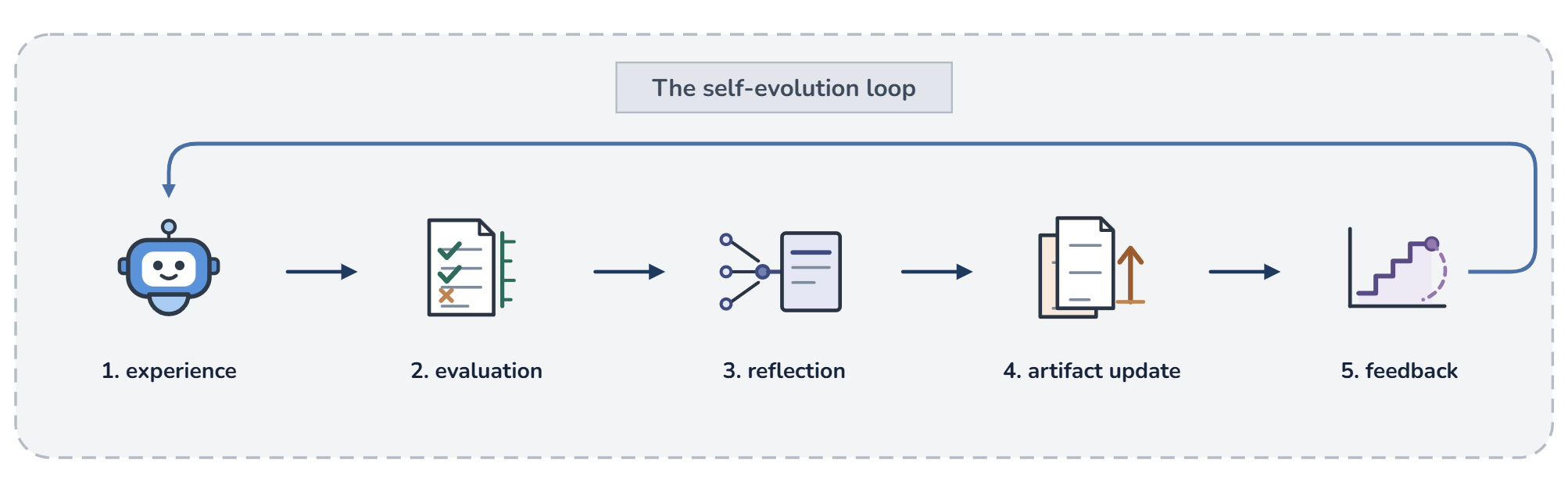}
    \caption{The self-evolution loop: experience, evaluation, reflection, artifact updates, and feedback.}
    \label{fig:self_evolution_loop}
\end{figure*}

In the self-evolution loop (Figure~\ref{fig:self_evolution_loop}), the system compresses part of its interaction trajectories into persistent artifacts and invokes them in later decisions. Compression makes experience reusable across tasks, but it also discards the state conditions, failure causes, and alternative actions of the original trajectory. We must therefore test whether the retained experience still produces task-relevant effects in new decisions. We define five properties along the stages an artifact passes through to become a capability:
\begin{enumerate}
\item \textbf{Artifact formation}: a persistent structure has formed and is connected to subsequent decision making;
\item \textbf{Behavioral activation}: under the same state and interface conditions, actions with and without the artifact differ by more than sampling noise;
\item \textbf{Task-relation realization}: the artifact produces a positive effect---the resulting actions better satisfy the relevant task relation overall, rather than merely differing from the original ones;
\item \textbf{Boundary-safe reuse}: the artifact does not undermine an established effect---in actual decision making, an established positive relation is not overturned by another part of the system;
\item \textbf{Capability extension}: a later version adds new positive effects while preserving existing ones.
\end{enumerate}

The dependency structure is $1 \rightarrow 2 \rightarrow (3 \text{ and } 4) \rightarrow 5$; we call this sequence the \emph{capability chain}. Properties 3 and 4 form a single \emph{checkpoint}: the former asks whether the artifact produces a positive effect, the latter whether it undermines an established one. When both hold, we say that \emph{capability formation} has occurred. Property 4 concerns only established positive relations that are overturned in actual decisions; gains and losses on fixed states already count toward Property 3. Property 5 is meaningful only for versions that pass the checkpoint. Diagnosis proceeds in this order: the first property that fails is the \emph{break}, and properties that depend on it are not judged.

The five properties apply at three scales. Properties 2 and 3 concern single moves and are judged on fixed states; Property 4 concerns established effects in actual decision making and is judged in complete games; Property 5 concerns evolution and is judged along the version lineage; and Property 1 is confirmed directly from call records. We call this \emph{three-scale diagnosis}, comprising state-level, game-level, and lineage-level diagnosis (Section~\ref{sec:design}). Game-level diagnosis does not judge wins and losses as such, but whether an established effect is overturned.

A \emph{task relation} is defined by the task objective: it specifies which actions advance the task in a given state, such as taking one's own immediate winning point, blocking the opponent's, or forming an open four. Environmental rules, such as not playing on an occupied point, are not task relations: every legal action satisfies them, yet they advance nothing. Property 3 therefore compares only legal actions; reducing rule violations does not by itself count as a positive effect.

\subsection{System-Level Unit of Analysis}\label{sec:system-unit}

Our object of diagnosis is the \emph{frozen system}: all artifacts that actually participate in decision making in a frozen version, taken as a whole. Decisions are produced jointly by several \emph{components}: skills propose candidates, a calling script decides which skill to query first and which answer to accept, and code branches override one another. A component may be correct on its own yet have its effect overturned by another once they are combined. With single components as the unit, a failure could only be blamed on ``a poorly written skill,'' leaving open whether the system as a whole has formed a capability. We therefore judge Properties 1--4 on the frozen system; which component a break lies in serves only as explanation and does not change the judgment. Formal definitions appear in Appendix~\ref{app:definitions}.

\begin{figure*}[!t]
    \centering
    \includegraphics[width=0.8\linewidth]{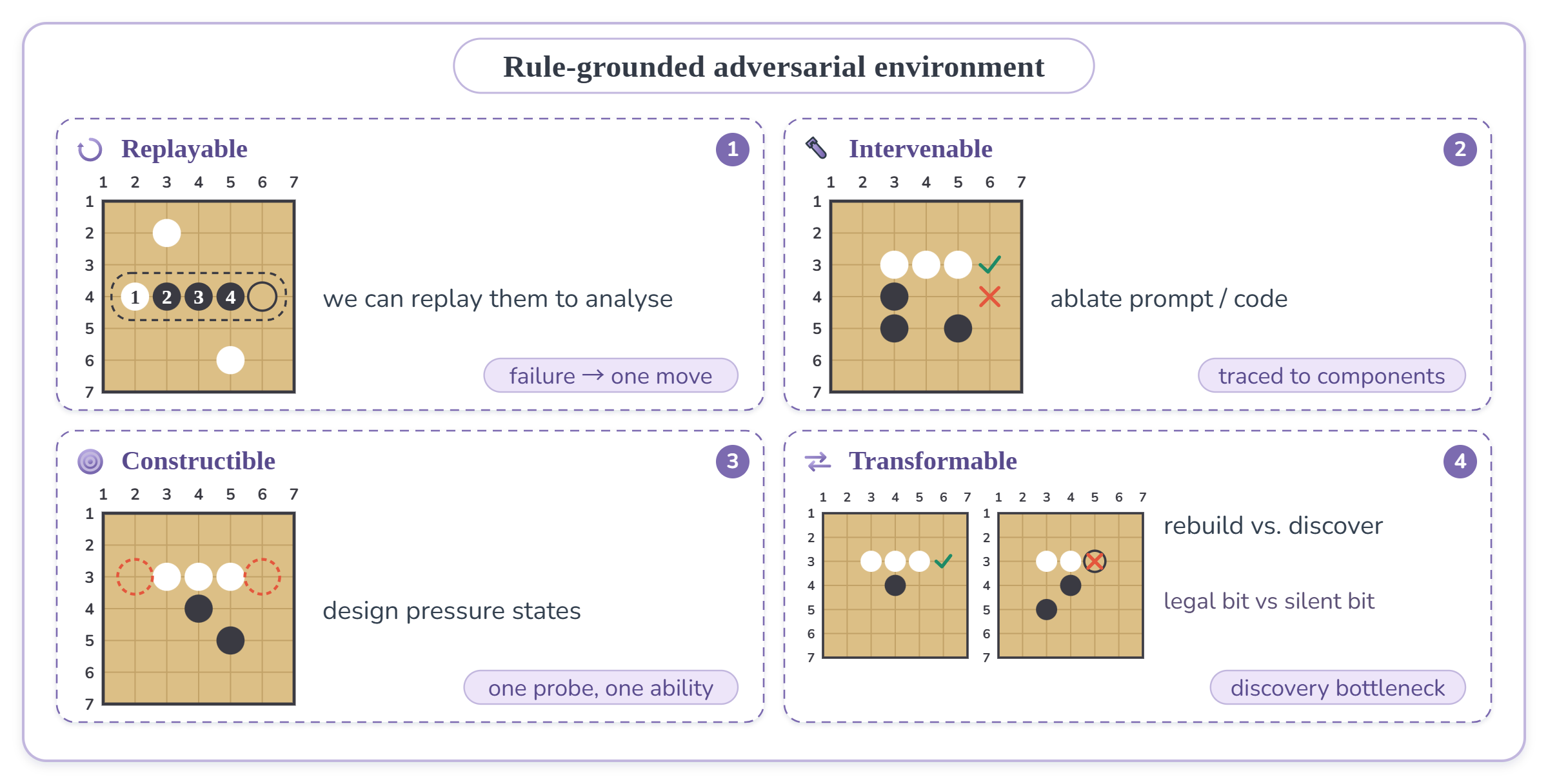}
    \caption{The three properties of the rule-based environment and the diagnostic scale each one supports: replayable states (state level), rule-judged task effects (state and game level), and fully recorded games and versions (game and lineage level).}
    \label{fig:rule_grounded_environment}
\end{figure*}

\section{Experimental Design}\label{sec:design}

The three mechanisms evolve autonomously with the same base model (DeepSeek-V4-Flash), environment, and opponents. After evolution, they are frozen and undergo three-scale diagnosis (Table~\ref{tab:experiments}). Judgments rely only on experimental data and rules.

\begin{table*}[!t]
\centering\small
\caption{Three-scale diagnosis and the properties it judges.}
\label{tab:experiments}
\begin{tabularx}{\textwidth}{>{\raggedright\arraybackslash}p{2.2cm}ZZ}
\toprule
Scale & Properties and criteria & Why this works \\
\midrule
State-level & Compare moves with and without the tested artifact on the same states. Property 2: move changes exceed sampling noise. Property 3: the pass rate among legal moves increases. & The states are identical and the tested artifact is the only difference, so any change in moves can only come from it. A rule program independently specifies the acceptable moves, distinguishing ``changed for the better'' from ``merely changed.'' \\
\addlinespace
Game-level & Play complete games between the frozen system and a programmatic opponent. Property 4: complete five at every immediate-win state. & At an immediate-win state, ``win in one move'' is already established and the rules specify the unique correct move. Failing to complete five shows that another part of the system overturned this effect, so no control condition is needed. \\
\addlinespace
Lineage-level & Compare adjacent versions state by state on the same states, along with their wins. Property 5: later versions retain prior effects while adding new ones; losing without adding, or barely changing behavior, fails. & Paired state-level comparison shows exactly what each version adds or loses, and whether its modifications still reach behavior. \\
\bottomrule
\end{tabularx}
\end{table*}

\subsection{Environment: Freestyle Gomoku}\label{sec:env}

Three properties of the environment make three-scale diagnosis possible (Figure~\ref{fig:rule_grounded_environment}). \emph{Replayable states}: any board can be restored exactly, so state-level diagnosis compares moves with and without an artifact on the same state, and set A can be expanded by rotation and reflection. \emph{Rule-judged task effects}: a rule program enumerates the acceptable moves of each state independently of the system under diagnosis and of any other model, so ``changed for the better'' is separated from ``merely changed'' for Properties 3 and 4. \emph{Fully recorded games and versions}: every move, call path, and version is archived, so game-level diagnosis can locate conflict states and lineage-level diagnosis can count newly passing and lost states.

The board is $15\times15$. Players alternate placing stones on empty intersections; playing on an occupied intersection or off the board is illegal. Forming a horizontal, vertical, or diagonal line of five stones (\emph{five in a row}) wins, and game length is counted in moves.

A point at which one move completes five is an \emph{immediate winning point}. A state in which it is the agent's turn and the agent has an immediate winning point is an \emph{immediate-win state}; if the opponent also has one, it is a \emph{conflict state}. Four in a row with both ends open is an \emph{open four}, and three in a row with both ends open is an \emph{open three}. In each state, the legal moves that satisfy the primary task relation form the acceptable set $G(s)$; a move in $G(s)$ \emph{passes}, and the fraction of states passed in a set is its \emph{pass rate}. All three opponents are programmatic (\texttt{rule-based}, \texttt{threat-aware}, and \texttt{search\_2}); the strongest, \texttt{search\_2}, searches two plies ahead. Appendix~\ref{app:gomoku-criteria} gives the full rules and criteria.

\begin{table*}[!t]
\centering\footnotesize
\caption{Signals and three-scale diagnosis results for the three mechanisms. Behavior change corresponds to Property 2; Property 1 is confirmed from call records for all three mechanisms (Section~\ref{sec:results}); Properties 3 and 4 must both hold.}
\label{tab:mechanism-summary}
\begin{tabularx}{\textwidth}{>{\raggedright\arraybackslash}p{1.4cm}>{\raggedright\arraybackslash}p{4.8cm}XXX>{\raggedright\arraybackslash}p{1.3cm}}
\toprule
 & Signals & \multicolumn{3}{c}{Three-scale diagnosis} & \\
\cmidrule(lr){2-2}\cmidrule(lr){3-5}
Mechanism & Accumulation / behavior change / metric gain & State-level: Property 3 & Game-level: Property 4 & Lineage-level: Property 5 & Break \\
\midrule
Reflexion & Six cells, seven memories each; moves change on 59.8\%--78.2\% of B states (noise: 46.7\%); overall pass rate rises from 8.0\% to 9.2\%--10.3\% on B and from 21.7\% to 22.5\%--27.5\% on A. & Legal pass rate improves in none of 12 combinations; \textbf{fails}. & Not judged. & Not judged. & Property 3 \\
\addlinespace
Voyager & Seven skills and a calling script; ablating components changes moves on 89--808 B states; open fours rise from 0/20 to 20/20. & \texttt{round1} $\rightarrow$ \texttt{full}: 459 newly passing and 73 lost on B; holds. & Completes five at 0/20 immediate-win states; \textbf{fails}. & Not judged. & Property 4 \\
\addlinespace
DGM & Six accepted descendants; OB-3 changes moves on at least 661 B states relative to P0; B pass rate rises from 48.7\% (P0) to 63.0\% (OB-7); evolutionary score rises from 150.4 (OB-1) to 312.6 (OB-7). & OB-3 vs.\ P0: 522 newly passing and 139 lost; holds. & Completes five at 19/19; holds. & OB-5 and OB-6 only lose; OB-7 barely changes behavior; \textbf{fails}. & Property 5 \\
\bottomrule
\end{tabularx}
\end{table*}

\subsection{Three Scales}

\textbf{State-level diagnosis.} We use 1,876 fixed states: set A contains 120 states obtained by rotating and reflecting manually constructed tactical states, and set B contains 1,756 states drawn from actual games during DGM evolution (construction in Appendix~\ref{app:gomoku-criteria}). The control removes the artifact under test: empty memory for Reflexion; \texttt{round1} for Voyager, which keeps only the first-round tactical skills and leaves the calling script unchanged; and the agent's initial strategy P0 for DGM. For Voyager and DGM, we thus test the effect of what evolution added. Because Reflexion samples its moves, we use the difference between two empty-memory runs as the noise baseline.

\textbf{Game-level diagnosis.} The frozen system plays 20 games against \texttt{search\_2}, each with a different random seed. We rebuild the board after every move, identify all immediate-win states, and record whether the system completes five; for games in which it does not, we keep the per-move call trace for explanation.

\textbf{Lineage-level diagnosis.} This applies only to mechanisms with a version lineage, here DGM. We judge Properties 1--4 on OB-3, the evolved version with the most wins against \texttt{search\_2}, and use lineage-level diagnosis to test whether later versions retain and extend its effects. Along the main lineage, we compare adjacent versions state by state on set B, count newly passing and lost states, and compare wins. We also build a differential pool of 7,234 states from archived evolution games and compare the moves of the last two versions, OB-6 and OB-7, on every state (Appendix~\ref{app:ob6-ob7-diff}).

\subsection{Three Mechanisms}
\begin{figure*}[!t]
    \centering
    \includegraphics[width=0.63\textwidth]{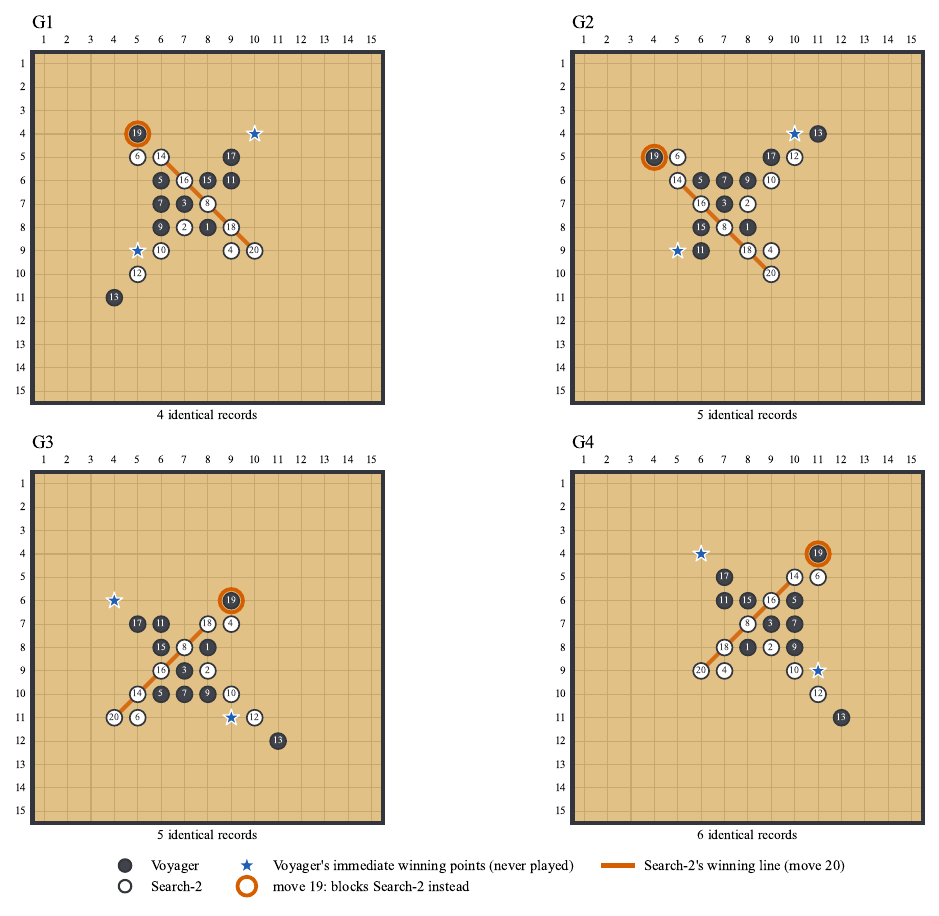}
    \caption{Voyager against \texttt{search\_2}. Twenty independent games with different random seeds converge to four move sequences (G1--G4); numbers give move order. In every sequence, Voyager forms an open four whose two immediate winning points (blue stars) stay empty. At move~19, a conflict state, \texttt{auto\_player} returns early on the defense branch and blocks one of \texttt{search\_2}'s immediate winning points (orange ring); \texttt{search\_2} completes five at the other on move~20 (orange line). Immediate winning points are recomputed from the rules; Appendix~\ref{app:voyager-call-order} lists them and the \texttt{auto\_player} call order.}
    \label{fig:voyager_search2_exact_trajectories}
\end{figure*}
The three mechanisms cover three artifact forms; implementation details are in Appendix~\ref{app:setup}. \textbf{Reflexion} writes game feedback into textual memories, split by opponent and feedback type (\texttt{outcome} gives only the result; \texttt{trace} adds the full move sequence) into six cells of seven entries each, all of which we test; the model chooses moves directly. \textbf{Voyager} writes experience as executable skills that only detect board features or propose candidates; for example, \texttt{detect\_immediate\_wins} finds immediate winning points, and \texttt{create\_open\_three} proposes attacking moves that form an open three. The final move is chosen by \texttt{auto\_player}, a calling script that the agent writes itself; it queries skills in a fixed order and returns as soon as one fires. \textbf{DGM} modifies strategy code directly, producing the lineage P0 $\rightarrow$ OB-1 $\rightarrow$ OB-2 $\rightarrow$ OB-3 $\rightarrow$ OB-5 $\rightarrow$ OB-6 $\rightarrow$ OB-7 (OB-4 is a side branch).

\section{Diagnostic Results}\label{sec:results}
\begin{figure}[!tb]
    \centering
    \includegraphics[width=\columnwidth]{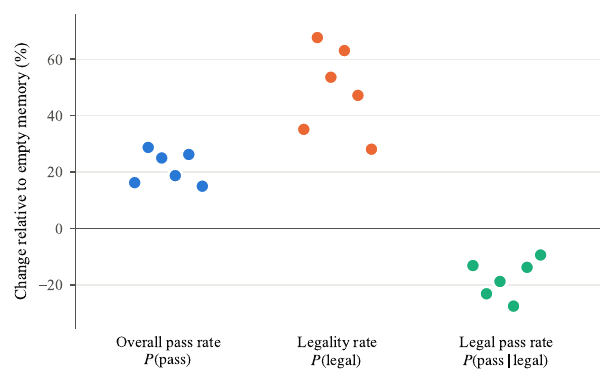}
    \caption{Reflexion on set B: change in each rate relative to empty memory, one dot per frozen memory (six in total; each memory keeps the same horizontal position in all three columns). All six raise the overall pass rate and the legality rate, yet all six lower the legal pass rate. The empty-memory rerun changes the three rates by $+2.5\%$, $+2.4\%$, and $+0.6\%$. Full results for both sets are in Table~\ref{tab:reflexion-calls-full}.}
    \label{fig:reflexion-legalization}
\end{figure}

Property 1 holds for all three mechanisms and is confirmed directly from call records: Reflexion's memories are included in every move prompt, Voyager's skill library and \texttt{auto\_player} are invoked at every move, and every DGM version supplies the \texttt{select\_move} function that the evaluator calls. We do not restate it below. Table~\ref{tab:mechanism-summary} summarizes the results: all three mechanisms show the three signals, yet none completes the capability chain, and each exhibits a different capability-formation break.

\subsection{Reflexion: Legalization Gain, Not Selection Gain}

After memories are loaded, moves change on 59.8\%--78.2\% of B states, above the noise baseline of 46.7\% (Property 2 holds), and the overall pass rate $P(\text{pass})$ rises on both sets (Table~\ref{tab:reflexion-calls-full}). Taken alone, this metric suggests that reflection helps the model play better.

By definition, however, $P(\text{pass}) = P(\text{legal}) \times P(\text{pass}\mid\text{legal})$, and the rise in overall pass rate comes entirely from a higher legality rate---a \emph{legalization gain} (Figure~\ref{fig:reflexion-legalization}). The legal pass rate, which measures move selection, improves in none of the 12 combinations and, on the larger, lower-noise set B falls by 1.6--4.4 percentage points. State-level pairing agrees: most newly passing states come from illegal moves becoming legal, and on states where both conditions yield legal moves, the net change is negative (Appendix~\ref{app:reflexion-statistics}). Memory suppresses rule violations but produces no \emph{selection gain}. \textbf{Task-relation realization (Property 3) fails}, and the break is at Property 3 (alternative explanations in Appendix~\ref{app:reflexion-alt}).
\begin{table*}[!t]
\centering\small
\caption{Wins against \texttt{search\_2}, B pass rate, and paired changes relative to the preceding version along the main DGM lineage.}
\label{tab:dgm-lineage}
\begin{tabularx}{\textwidth}{lYYY}
\toprule
Version & Wins & B pass rate & Newly passing / lost \\
\midrule
P0 & 20/20 & 48.7\% & --- \\
OB-3 & 19/20 & 70.5\% & $+522$ / $-139$ (vs.\ P0) \\
OB-5 & 9/20 & 64.6\% & 0 / $-103$ \\
OB-6 & 0/20 & 63.0\% & 0 / $-29$ \\
OB-7 & 0/20 & 63.0\% & 1 / 0 \\
\bottomrule
\end{tabularx}
\end{table*}
\subsection{Voyager: Compositional Override}\label{sec:case-open-three}

Ablating any component changes moves on 89 to 808 B states (Property 2 holds), and from \texttt{round1} to \texttt{full}, 459 B states become passing and 73 are lost (Property 3 holds; ablation data in Appendix~\ref{app:voyager-trajectories}). Open fours in games also rise from 0/20 to 20/20.

In game-level diagnosis, however, Voyager reaches an immediate-win state in all 20 games yet completes five at none of them (0/20), blocking the opponent's immediate winning point instead and losing every game (Figure~\ref{fig:voyager_search2_exact_trajectories}). All 20 are conflict states. Call traces locate the break in how the components combine. The ``win if possible'' logic in \texttt{evaluate\_threats} is correct on its own; the defense-first rule in \texttt{auto\_player} is reasonable when only the opponent has an immediate winning point; and \texttt{create\_open\_three} drives attacks that form open fours. These attacks, however, make conflict states frequent, and at such states \texttt{auto\_player} returns at its first branch, which blocks the opponent (defense-first early return), so neither its own immediate-win check nor the ``win if possible'' logic in \texttt{evaluate\_threats} is ever reached (Appendix~\ref{app:voyager-call-order}). Each component is sound on its own, yet in combination they overturn an established effect---a \emph{compositional override}. \textbf{Boundary-safe reuse (Property 4) fails}, and the break is at Property 4 (alternative explanations in Appendix~\ref{app:voyager-alt}).

\subsection{DGM: Loss-Only Drift After Capability Formation}

DGM shows all three signals (Table~\ref{tab:mechanism-summary}). OB-3, the evolved version with the most wins, passes the checkpoint. Relative to P0, 522 B states become passing and 139 are lost, a net gain of 383 (Properties 2 and 3 hold). It completes five at all 19 immediate-win states in its 20 games (19/19; Property 4 holds) and wins 19 games, one fewer than P0. OB-3 has formed a capability.

Lineage-level diagnosis, however, shows no extension afterward (Table~\ref{tab:dgm-lineage}). OB-5 and OB-6 add no newly passing states and only lose, while wins fall to 0---\emph{loss-only drift}. OB-7 differs from OB-6 on only one of the 7,234 states in the differential pool, so its modification barely reaches actual decisions (Appendix~\ref{app:ob6-ob7-diff}). Meanwhile, the evolutionary score still rises from OB-3 to OB-5 (Appendix~\ref{app:dgm-alt}). \textbf{Capability extension (Property 5) fails}, and the break is at Property 5 (alternative explanations in Appendix~\ref{app:dgm-alt}).

\section{Discussion}
Although derived in Gomoku, the properties concern task relations rather than the game; we conjecture, without having tested it, that the same breaks arise wherever artifacts are reused through a fixed interface. Consider a GUI agent with two skills. Skill A moves the cursor to a target element and clicks; it can reach the target (a positive effect), land on a non-interactive region (an ineffective change), or open an unrelated page (a harmful change), and a version that only repairs malformed coordinates would yield a legalization gain. Skill B, written for pages behind a pop-up, first clicks to dismiss the pop-up and then moves to its own target. Each skill reaches its target alone, but when B runs after A, its first click hits whatever lies under the cursor on the page A just reached and navigates away: a compositional override of A's established effect. Testing this requires replayable screens and a rule-based checker of application state (Section~\ref{sec:env}).

\section{Conclusion}

Most work supports self-evolving methods with artifact accumulation, behavioral change, and metric gains---signal-based evaluation---without defining what counts as capability formation. We define the transition from artifact to capability as five properties and judge them with three-scale diagnosis. Under the same base model, environment, and opponents, the diagnosis detects three capability-formation breaks: a legalization gain in Reflexion, a compositional override in Voyager, and loss-only drift in DGM. These counterexamples show that the three signals are useful but insufficient to determine whether a capability has formed. Evaluating the benefits of self-evolution requires directly testing positive effects, the preservation of established effects, and extension across versions, at the state, game, and lineage levels.

\section*{Limitations}

How the three break types are distributed across mechanisms and tasks requires diagnosis at a larger scale. We cover three representative self-evolving mechanisms, whose artifacts are textual memories, executable skill libraries, and code-version lineages, but not every form of self-evolution: multi-agent co-evolution or direct parameter updates may break at other stages. Which properties can be judged also depends on a mechanism's structure; for example, capability extension can be judged only for mechanisms with a version lineage. During diagnosis, Reflexion chooses moves with the model directly, whereas Voyager and DGM use evolved code. Whether different base models, longer evolution, or repeated runs would move the breaks elsewhere remains to be tested.

Finally, our empirical scope is limited to freestyle Gomoku. We chose it because rules can independently judge whether a move is correct and states can be replayed exactly; this also means that its task relations are local and enumerable. Task relations in open-ended tasks are often hard to determine from rules, and how to operationalize the framework there remains open.

\FloatBarrier
\bibliography{custom}

\appendix
\renewcommand{\thetable}{\thesection.\arabic{table}}

\FloatBarrier
\section{Diagnostic Definitions}\label{app:definitions}

Let $s$ be a fixed state, let $\pi_A(s)$ denote the move selected by the system with the artifact under comparison loaded, and let $\pi_\varnothing(s)$ denote the move under the control condition. Let $G(s)$ be the set of legal moves in $s$ that satisfy the relevant task relation; a move is counted as passing if and only if it belongs to $G(s)$.

\paragraph{Behavioral activation.} We count the states for which $\pi_A(s)\neq\pi_\varnothing(s)$. For mechanisms in which the model samples moves, we compare this count with the same quantity computed between two independent control runs (the noise baseline); a change counts as activation only if it exceeds that baseline. Voyager and DGM select moves deterministically in code, so any changed output counts as activation.

\paragraph{Task-relation realization.} At the call level, the pass rate decomposes as $P(\text{pass})=P(\text{legal})\times P(\text{pass}\mid\text{legal})$; we judge improvement only by whether $P(\text{pass}\mid\text{legal})$ increases. At the state level, newly passing states are classified by source as ``illegal $\rightarrow$ pass'' or ``legal non-pass $\rightarrow$ pass.'' The former indicates only that a move became legal and does not count as a positive task effect.

\paragraph{Boundary-safe reuse.} We reconstruct each game move by move and identify every immediate-win state. If there are $n$ such states and the system completes five in $k$ of them, we report $k/n$; any $k<n$ indicates that an already-established winning relation was overturned.

\paragraph{Capability extension.} Let $C_v$ be the set of states passed by version $v$ on set B. Between adjacent versions, newly passing and lost states are $N=C_{v'}\setminus C_v$ and $L=C_v\setminus C_{v'}$, respectively; we also compare wins against \texttt{search\_2}. This criterion applies only to iterations after the version that passes the checkpoint. Such an iteration fails to preserve and extend capability when $|N|=0$ and $|L|>0$, or when the number of wins declines. We classify an iteration as stagnant when $|N|+|L|$ is close to zero and the selected moves are almost unchanged.

\FloatBarrier
\section{Environment and Fixed States}\label{app:gomoku-criteria}

An independent rule-based referee checks the legality and outcome of every move. When the agent has an immediate winning point and it is its turn, the uniquely correct move under the rules is to complete five, even if the opponent also has an immediate winning point: completing five ends the game immediately, whereas blocking only allows the game to continue.

We construct $G(s)$ as follows. Set A is formed from 22 manually constructed tactical states through rotations and reflections; we retain 120 states with unambiguous task relations. Set B consists of 1,756 deduplicated states sampled from games during DGM evolution and balanced across task-relation categories. For both sets, a rule program enumerates all legal moves satisfying the primary task relation in each state; for example, the immediate-win relation includes every move that completes five at once.

In game-level diagnosis, each system or version plays 20 complete games against \texttt{search\_2}, with a limit of 200 moves per game. Games exceeding the limit are recorded as undecided.

\FloatBarrier
\section{Reflexion Diagnostic Data}\label{app:reflexion-statistics}
\setcounter{table}{0}

We evaluate seven diagnostic conditions: an empty-memory condition and six frozen-memory conditions. The six memories correspond to the \texttt{rule-based}, \texttt{threat-aware}, and \texttt{search\_2} opponent cells under the \texttt{outcome} and \texttt{trace} feedback settings. Each memory contains seven entries. For each state and condition, we query the model once with temperature 0.2 and \texttt{max\_tokens}=512. We independently run the empty-memory condition a second time to estimate sampling noise.

\begin{table*}[!t]
\centering\small
\caption{Reflexion call outcomes on sets A and B. Each state is queried once under each condition. The opponent cell associated with each memory condition is shown in parentheses.}
\label{tab:reflexion-calls-full}
\begin{tabularx}{\textwidth}{ZcYYYY}
\toprule
Condition & Set & Legality rate & \# Pass & $P(\text{pass})$ & $P(\text{pass}\mid\text{legal})$ \\
\midrule
Empty memory & A & 58.3\% & 26 & 21.7\% & 37.1\% \\
Empty memory & B & 49.8\% & 140 & 8.0\% & 16.0\% \\
\addlinespace
Empty memory (second run) & A & 65.8\% & 26 & 21.7\% & 32.9\% \\
Empty memory (second run) & B & 51.0\% & 144 & 8.2\% & 16.1\% \\
\addlinespace
\texttt{outcome} (\texttt{threat-aware}) & A & 75.8\% & 33 & 27.5\% & 36.3\% \\
\texttt{outcome} (\texttt{threat-aware}) & B & 67.3\% & 164 & 9.3\% & 13.9\% \\
\addlinespace
\texttt{trace} (\texttt{threat-aware}) & A & 96.7\% & 29 & 24.2\% & 25.0\% \\
\texttt{trace} (\texttt{threat-aware}) & B & 83.5\% & 180 & 10.3\% & 12.3\% \\
\addlinespace
\texttt{outcome} (\texttt{rule-based}) & A & 76.7\% & 31 & 25.8\% & 33.7\% \\
\texttt{outcome} (\texttt{rule-based}) & B & 76.5\% & 175 & 10.0\% & 13.0\% \\
\addlinespace
\texttt{trace} (\texttt{rule-based}) & A & 93.3\% & 27 & 22.5\% & 24.1\% \\
\texttt{trace} (\texttt{rule-based}) & B & 81.2\% & 166 & 9.5\% & 11.6\% \\
\addlinespace
\texttt{outcome} (\texttt{search\_2}) & A & 83.3\% & 32 & 26.7\% & 32.0\% \\
\texttt{outcome} (\texttt{search\_2}) & B & 73.3\% & 178 & 10.1\% & 13.8\% \\
\addlinespace
\texttt{trace} (\texttt{search\_2}) & A & 75.8\% & 30 & 25.0\% & 33.0\% \\
\texttt{trace} (\texttt{search\_2}) & B & 63.8\% & 162 & 9.2\% & 14.5\% \\
\bottomrule
\end{tabularx}
\end{table*}

\begin{table*}[!t]
\centering\small
\caption{Results for the six frozen memories on set B ($n=1{,}756$). Net changes and $p$-values are based on state-level paired outcomes using the exact McNemar test.}
\label{tab:reflexion-b-cells}
\begin{tabularx}{\textwidth}{>{\raggedright\arraybackslash}p{3.4cm}YYYY}
\toprule
Condition & Move change & Net change & $p$ & $P(\text{pass}\mid\text{legal})$ \\
\midrule
Empty memory (second run; noise baseline) & 46.7\% & $+4$ & 0.73 & 16.0\% $\rightarrow$ 16.1\% \\
\texttt{threat-aware}, \texttt{outcome} & 61.3\% & $+24$ & 0.0433 & 13.9\% \\
\texttt{threat-aware}, \texttt{trace} & 78.2\% & $+40$ & 0.0037 & 12.3\% \\
\texttt{rule-based}, \texttt{outcome} & 62.9\% & $+35$ & 0.0033 & 13.0\% \\
\texttt{rule-based}, \texttt{trace} & 72.8\% & $+26$ & 0.0478 & 11.6\% \\
\texttt{search\_2}, \texttt{outcome} & 65.5\% & $+38$ & 0.0031 & 13.8\% \\
\texttt{search\_2}, \texttt{trace} & 59.8\% & $+22$ & 0.0548 & 14.5\% \\
\bottomrule
\end{tabularx}
\end{table*}

\begin{table*}[!t]
\centering\small
\caption{State-level paired outcomes for the \texttt{threat-aware} memories and the sources of newly passing states. ``Net change among legal moves'' is computed only over states on which the model makes a legal move both with empty memory and with loaded memory; it equals the number of ``legal non-pass $\rightarrow$ pass'' transitions minus the number of ``pass $\rightarrow$ legal non-pass'' transitions.}
\label{tab:reflexion-pairs}
\begin{tabularx}{\textwidth}{cc*{6}{Y}}
\toprule
Feedback & Set & Newly passing & Lost & $p$ & Illegal $\rightarrow$ pass & Legal non-pass $\rightarrow$ pass & Net change among legal moves \\
\midrule
\texttt{outcome} & A & 8 & 1 & 0.039 & 6 & 2 & $+1$ \\
\texttt{outcome} & B & 77 & 53 & 0.043 & 53 & 24 & $-12$ \\
\addlinespace
\texttt{trace} & A & 7 & 4 & 0.55 & 4 & 3 & $-1$ \\
\texttt{trace} & B & 111 & 71 & 0.0037 & 72 & 39 & $-19$ \\
\bottomrule
\end{tabularx}
\end{table*}

\subsection{Alternative Explanations}\label{app:reflexion-alt}

\paragraph{Move changes are merely sampling noise.}
Between the two empty-memory runs, the selected move differs on 46.7\% of states in set B. Passes change in both directions at similar rates (41 versus 37), suggesting little net change. With the six memories loaded, moves change on 59.8\%--78.2\% of states, and every condition yields a positive net change (Table~\ref{tab:reflexion-b-cells}), indicating that the memory-induced changes exceed the observed sampling variation.

\paragraph{More legal moves dilute the legal pass rate.}
Newly legal moves may occur on more difficult states, thereby lowering the legal pass rate. We address this possibility by considering only states for which both the empty-memory and loaded-memory conditions produce legal moves. On set B, the net change within this legal subset is $-12$ and $-19$ for the \texttt{outcome} and \texttt{trace} conditions, respectively (Table~\ref{tab:reflexion-pairs}). Thus, although the number of states passing the diagnostic increases, most of the increase comes from illegal actions becoming legal rather than from improvements among actions that were already legal.

\paragraph{The memories may not match the diagnostic states.}
The memories are organized into three opponent-specific groups, whereas the diagnostic states do not encode opponent identity. Any memory may therefore be irrelevant to some of the states. We consequently evaluate all six memories separately. Across the 12 combinations of memory condition and state set (A or B), the legal pass rate improves in none of them (Table~\ref{tab:reflexion-calls-full}), so the conclusion does not depend on which memory is used. Because set A contains relatively few states, its noise-induced fluctuation is $-4.2$ percentage points; the main evidence therefore comes from set B, where the corresponding fluctuation is only $+0.1$ percentage points.

\FloatBarrier
\section{DGM Diagnostic Data}\label{app:dgm-records}
\setcounter{table}{0}

During evolution, the agent autonomously decides whether to retain each candidate and which version should serve as the parent for the next generation based on the evaluation results. The main lineage is P0 $\rightarrow$ OB-1 $\rightarrow$ OB-2 $\rightarrow$ OB-3 $\rightarrow$ OB-5 $\rightarrow$ OB-6 $\rightarrow$ OB-7; OB-4 is a branch derived from OB-3.

\begin{table*}[!t]
\centering\small
\caption{Terminal outcomes against \texttt{search\_2} and pass rates on sets A and B for the main DGM lineage. The denominators for A and B are 120 and 1,756, respectively.}
\label{tab:dgm-full}
\begin{tabularx}{\textwidth}{ZYYY}
\toprule
Version & \texttt{search\_2} wins/undecided/losses & A & B \\
\midrule
P0 & 20/0/0 & 66.7\% & 48.7\% \\
OB-1 & 0/0/20 & 80.0\% & 78.6\% \\
OB-2 & 15/0/5 & 74.2\% & 70.7\% \\
OB-3 & 19/0/1 & 80.0\% & 70.5\% \\
OB-5 & 9/11/0 & 66.7\% & 64.6\% \\
OB-6 & 0/11/9 & 66.7\% & 63.0\% \\
OB-7 & 0/11/9 & 66.7\% & 63.0\% \\
\bottomrule
\end{tabularx}
\end{table*}

\paragraph{OB-6/OB-7 differential pool.}\label{app:ob6-ob7-diff}
The differential pool was constructed from 472 archived games collected during DGM evolution. After exact deduplication by board state and player to move, it contained 7,212 states. We then added the 22 tactical states before expansion into set A, yielding 7,234 states in total. We queried the move-selection function of OB-6 and OB-7 separately on every state. Their selected moves differed on only one state, consistent with the fact that OB-7 adds only one newly passing state on set B.

\subsection{Alternative Explanations}\label{app:dgm-alt}

\paragraph{Using OB-3 as the reference is cherry-picking.}
After OB-3, both OB-5 and OB-6 add zero newly passing states on set B. The issue is therefore not simply that later versions failed to return to the peak; they added no new passing states at all, regardless of the reference version. Using OB-1, which achieves the highest pass rate, as the reference leads to the same conclusion: the B pass rate only decreases thereafter (Table~\ref{tab:dgm-full}).

\paragraph{Local misalignment between the score and the diagnostic results.}
The composite score and the diagnostic results show broadly similar trends across versions: from OB-1 to OB-3, both the score and the number of wins against \texttt{search\_2} increase. However, a local mismatch appears from OB-3 to OB-5: the score increases from 347.0 to 357.8, whereas the number of wins decreases from 19 to 9. This occurs because the game-length term in the evaluation function compensates for the decline in the win term. Thus, the composite score broadly reflects changes in version performance, but this local mismatch shows that a higher score alone does not guarantee improved task capability. Whether later versions preserve or extend existing capabilities must therefore be assessed using both fixed-state diagnostics and complete-game outcomes.

In addition, the score remains unchanged from OB-6 to OB-7, while only one move differs across the 7,234 replayed states. At the same time, OB-6's B pass rate (63.0\%) still exceeds P0's (48.7\%). The simultaneous occurrence of these independent state-level results and score stagnation indicates that the later stagnation cannot be attributed solely to the evaluation function; rather, the code changes barely entered the actual decision path.

\paragraph{Set B is derived from DGM's own games and may favor particular versions.}
The direction of change on set A is consistent with that on set B: after OB-3, the A pass rate decreases from 80.0\% to 66.7\% (Table~\ref{tab:dgm-full}).

\FloatBarrier
\section{Voyager Diagnostic Data}\label{app:voyager-trajectories}
\setcounter{table}{0}

For each ablation, we copied the frozen configuration and removed the specified component while keeping the remaining components and \texttt{auto\_player} unchanged. \texttt{round1} was implemented by removing \texttt{evaluate\_board} and the components that depend on it.

\begin{table*}[!t]
\centering\small
\caption{Game outcomes against \texttt{search\_2} for the Voyager configurations (20 games per configuration).}
\label{tab:voyager-ablation-games}
\begin{tabularx}{\textwidth}{ZYYY}
\toprule
Configuration & Own open four formed & Wins & Mean length (moves) \\
\midrule
\texttt{full} & 20/20 & 0/20 & 20.0 \\
\texttt{-open3} (without \texttt{create\_open\_three}) & 20/20 & 20/20 & 129.0 \\
\texttt{-minimax} (without \texttt{minimax\_search}) & 0/20 & 0/20 & 20.0 \\
\texttt{round1} (first-round tactical skills only) & 0/20 & 0/20 & 18.0 \\
\bottomrule
\end{tabularx}
\end{table*}

\begin{table*}[!t]
\centering\small
\caption{Pass counts for the Voyager configurations on sets A and B, with paired changes relative to the full system (moves changed / newly passing / lost).}
\label{tab:voyager-ablation-static}
\begin{tabularx}{\textwidth}{ZYY}
\toprule
Configuration & A (/120) & B (/1,756) \\
\midrule
\texttt{full} & 88 & 1,076 \\
\texttt{-open3} (without \texttt{create\_open\_three}) & 96; 8/8/0 & 1,345; 330/274/5 \\
\texttt{-minimax} (without \texttt{minimax\_search}) & 88; 0/0/0 & 1,084; 89/18/10 \\
\texttt{round1} (first-round tactical skills only) & 80; 16/0/8 & 690; 808/73/459 \\
\bottomrule
\end{tabularx}
\end{table*}

\subsection{Conflict States and Call Order}\label{app:voyager-call-order}

We replay all 20 games and recompute immediate winning points from the rules. Each game has exactly one immediate-win state for Voyager, at move~19, and every one is a conflict state: Voyager holds an open four with two empty immediate winning points, and \texttt{search\_2} also has two. In all 20 games Voyager plays on one of \texttt{search\_2}'s points, and \texttt{search\_2} completes five at the other on move~20. Table~\ref{tab:voyager-conflict} lists the four groups; coordinates are one-based (row, column), as in Figure~\ref{fig:voyager_search2_exact_trajectories}.

\begin{table*}[!t]
\centering\small
\caption{The conflict state at move~19 in each Voyager trajectory group against \texttt{search\_2}. Immediate winning points are recomputed from the rules.}
\label{tab:voyager-conflict}
\begin{tabularx}{\textwidth}{lcYYY}
\toprule
Group & Records & Voyager's immediate winning points & \texttt{search\_2}'s immediate winning points & Move played \\
\midrule
G1 & 4 & (4,10), (9,5) & (4,5), (9,10) & (4,5) \\
G2 & 5 & (4,10), (9,5) & (5,4), (10,9) & (5,4) \\
G3 & 5 & (6,4), (11,9) & (6,9), (11,4) & (6,9) \\
G4 & 6 & (4,6), (9,11) & (4,11), (9,6) & (4,11) \\
\bottomrule
\end{tabularx}
\end{table*}

The order of checks in the frozen \texttt{auto\_player} explains this outcome. Its first branch checks the opponent's immediate winning points and returns as soon as one exists; the check for the agent's own immediate win comes second and is therefore never reached in a conflict state:
{\footnotesize
\begin{verbatim}
# 0. Check opponent immediate wins FIRST
opp_wins = detect_immediate_wins(board, opponent)
if opp_wins:
    row, col = opp_wins[0]
    write_move(...); return
# 1. Check my immediate win
my_wins = detect_immediate_wins(board, agent_color)
if my_wins:
    row, col = my_wins[0]
    write_move(...); return
\end{verbatim}
}
Each branch is reasonable on its own: blocking is correct when only the opponent can win, and completing five is correct when only the agent can. The failure arises only when \texttt{create\_open\_three} drives the game into states where both apply.

\subsection{Alternative Explanations}\label{app:voyager-alt}

\paragraph{A single ineffective component cannot explain the failure.}
Holding all other skills and the calling script fixed, removing \texttt{create\_open\_three} still allows the system to form an open four in all 20 games and win all 20 games, whereas the full system wins 0/20 games. This intervention changes the subsequent game trajectories, but that trajectory change is part of the total effect of removing \texttt{create\_open\_three} from the full system. The results show that, with \texttt{auto\_player} and the remaining skills retained, the system can still form an open four and win. The call traces further show that prioritizing the own-immediate-win branch in conflict states would also prevent the same failure path. Thus, the problem is not that \texttt{create\_open\_three} or \texttt{auto\_player} fails to provide a positive contribution in isolation. Rather, their combination uses a call order that does not correctly handle conflict states.

\FloatBarrier
\section{Experimental Setup}\label{app:setup}\label{app:implementation}
\setcounter{table}{0}

All three mechanisms use DeepSeek-V4-Flash as the base model and evolve through games against the same three programmatic opponents: \texttt{rule-based}, \texttt{threat-aware}, and \texttt{search\_2}. We do not intervene in the artifacts during evolution, and we freeze them during diagnosis. The descriptions below specify the implementations used in our experiments; they preserve the core update paradigm of each mechanism.

\textbf{Reflexion.}
Each opponent has a separate memory cell. After each game, the system generates one textual lesson from the game feedback and appends it to the corresponding cell. Each cell retains at most seven memory entries. For diagnosis, we use the frozen state after the seventh entry has been written. The \texttt{outcome} condition provides only the win/loss result, whereas the \texttt{trace} condition also provides the complete move sequence. At decision time, all entries in the current cell are included in the move prompt, and the model directly outputs an action. Our implementation does not retrieve, filter, or weight memories based on the current state.

\textbf{Voyager.}
The agent writes its experience as executable Python skill files. The frozen configuration contains seven agent-written files: seven skills and the calling script \texttt{auto\_player}. Our diagnostic target is the complete decision system comprising the skill files and \texttt{auto\_player}. Skills detect board features or propose candidate moves, but do not directly output the final action. \texttt{auto\_player} invokes the skills in a fixed order and selects the move according to the branch that fires. During evolution, the agent modifies existing skills or adds new ones based on the previous round's evaluation, then enters the next round with the updated skill library.

\textbf{DGM.}
At each round, the agent modifies the strategy source code to produce a candidate version. Every version exposes a \texttt{select\_move(board, player)} interface, which the evaluator calls directly to obtain a move. Each candidate plays $5$, $10$, and $5$ games against the \texttt{rule-based}, \texttt{threat-aware}, and \texttt{search\_2} opponents, respectively. A composite evaluation score determines whether the candidate is retained and which version is used as the parent in the next round; candidates that are not selected remain in the version archive. This process produces a code lineage from the initial strategy P0 to later versions.

Lineage-level diagnosis uses OB-3 as the reference because it records the largest number of wins against \texttt{search\_2} during evolution. The evaluation score is computed programmatically and serves as a reference signal for the agent's retention decisions. For each candidate, the score is computed from $5$, $10$, and $5$ games against \texttt{rule-based}, \texttt{threat-aware}, and \texttt{search\_2}, respectively:
\begin{align}
S={}&50\,\mathrm{WR}_{\mathrm{RB}}+100\,\mathrm{WR}_{\mathrm{TA}}+150\,\mathrm{WR}_{\mathrm{SD2}}\\
&+20\max_o B_o+\overline{L}_{\mathrm{SD2}}+2\sum_o F_o,
\end{align}
where $\mathrm{WR}$ is the win rate against the corresponding opponent, $B_o$ is the block rate against opponent $o$: among the agent's moves made when $o$ has an immediate winning point, the fraction that occupy one of those points, $\overline{L}_{\mathrm{SD2}}$ is the average game length (in moves) against \texttt{search\_2}, and $F_o$ is the number of open fours formed against opponent $o$. The scores of the main-line versions are 150.4 for OB-1, 275.2 for OB-2, 347.0 for OB-3, 357.8 for OB-5, 312.6 for OB-6, and 312.6 for OB-7. OB-5 scores higher than OB-3 despite winning fewer games against \texttt{search\_2}: the game-length term increases as games are prolonged and offsets the decline in the win-rate term.

\FloatBarrier
\section{Extended Related Work}\label{app:related}

This appendix supplements the discussion with 20 related works; Table~\ref{tab:related-compare} compares their design elements with ours.

\subsection{Process Evaluation and Version Comparison}

\begin{itemize}
  \item \textbf{BenchTrace}~\citep{huang2026benchtrace} truncates a test snapshot before the target failure occurs, lets each framework and the ReAct baseline continue from the same prefix, and uses a rule-based detector to determine whether the failure is reproduced. It thereby measures failure-avoidance rate and also evaluates whether reflection can localize and diagnose failures. On Qwen3-32B and GPT-4.1, the end-to-end pass rate of reflection under this funnel-style, stage-wise evaluation is below 30\%.
  \item \textbf{EvoAgentBench}~\citep{gao2026evoagentbench} uses the accuracy gain over Vanilla for each ``task $\times$ evolutionary state'' cell and infers bottlenecks in experience encoding, routing, and adoption from the resulting gain patterns.
  \item \textbf{SEA-Eval}~\citep{jiang2026seaeval} measures token consumption and success rate at each position in task sequences of length 5, and distinguishes genuine evolution from pseudo-evolution according to whether the cost converges.
  \item \textbf{Huxley-G\"odel Machine}~\citep{wang2026hgm} estimates the meta-productivity of nodes in a self-modifying tree using an aggregate of descendant performance (CMP). On SWE-Verified-60, its estimator has a weighted correlation of 0.778 with empirical CMP, compared with 0.285 for DGM.
  \item \textbf{SAGE}~\citep{wang2026sage} compares the incumbent skill with a candidate edit on the same validation set. It retains only items on which the two versions disagree and applies a one-sided exact conditional binomial test (equivalent to a one-sided exact McNemar test when $\lambda=1$) as a statistical admission criterion for the edit.
\end{itemize}

Our version-by-version comparison in the DGM experiment is similar to SAGE in that both compare individual items in a fixed set. The comparison target here is the rule-defined state-level pass rate.

\subsection{Memory and Reflection Artifacts}

\begin{itemize}
  \item \textbf{\citet{zhang2026usefulmem}} compares different memory-integration strategies and finds that the utility of text memories continuously integrated by an LLM first increases and then decreases. On an ARC-AGI slice containing 19 problems that GPT-5.4 solved completely without memory, accuracy still falls to 52.6\% by the tenth round of streaming integration, even when the standard solution is provided at every step.
  \item \textbf{\citet{dixit2026honest}} analyzes publicly available Reflexion logs. In 16 ALFWorld environments with a reflection near-duplicate rate of at least 0.5, none of the 121 reflections mentions the correct object. Their no-memory control was run separately, with a different trial budget from the original logs.
  \item \textbf{ExpWeaver}~\citep{zhao2026expweaver} compares different timings for incorporating experience into decision making: not using it, using it only at initialization, using it at every step, and triggering it on demand. Its causal ablation either blanks out the retrieved content at the point where the agent triggers retrieval or randomly injects experience at the average trigger frequency. The outcome is measured by success over the entire trajectory. On Qwen3-32B, G-Memory, and ALFWorld, ExpWeaver achieves 83.33, compared with 75.37 when the retrieved content is blanked.
  \item \textbf{\citet{xiong2025experience}} finds that the more similar retrieved memories are to the current input, the more closely the agent's output resembles the output recorded in memory. They also construct an ``error-free'' variant that keeps the retrieved examples unchanged but replaces the recorded execution with a correct execution, allowing them to compare the effect of memory-content quality.
\end{itemize}

These works measure memory effects at the task, trial, or trajectory level. We instead determine the direction of memory-induced single-step action changes on fixed states.

\subsection{Attributing the Effects of Skill and Code Artifacts}

\begin{itemize}
  \item \textbf{ASSAY}~\citep{wang2026assay} randomly masks skills on software-development tasks and estimates each skill's effect as the mean difference in binary task outcomes for each ``skill $\times$ task'' pair. With GPT-5.1, 103 skills, and 15 development tasks, more than 90\% of skills have a cross-task effect range greater than 0.40.
  \item \textbf{\citet{hu2026skilltheater}} creates variants of a skill's meaning, wording, name, and content in single-turn question answering and pairs them with a no-skill condition. It measures answer flips and the consistency of self-reported attribution. On mathematical tasks, attribution consistency does not exceed 0.43 across models and conditions.
  \item \textbf{\citet{dong2026skillsharmful}} compares full runs of the same task with and without a skill and manually annotates failure causes. Of 125 functional failures, 68.8\% are attributed to task implementation errors.
  \item \textbf{Counterfactual Trace Auditing}~\citep{zhou2026cta} uses dynamic time warping (DTW) to align stages in two independent trajectories, with and without a skill, and compares patterns of behavioral influence. The authors describe this comparison as descriptive rather than a causal estimand.
  \item \textbf{Skill Following}~\citep{cho2026skillfollowing} computes the outcome difference between executions with and without a skill, restricted to tasks in which the agent actually retrieved the skill. On Math500, Llama-3.3-70B has an aggregate improvement of $+14.2$ percentage points, whereas this metric is $-39.4$ percentage points.
  \item \textbf{Library Drift}~\citep{zhang2026librarydrift} enables or disables different skill-library governance mechanisms at the full-run level and tallies each skill's successes and failures in injection trials.
  \item \textbf{\citet{pelleriti2026evolvecoding}} analyzes the types of code edits along an evolutionary lineage and replays saved generation prompts. About 30\% of added lines exactly match lines removed earlier; after replaying 36 breakthrough events, the median score ratio is 0.76.
\end{itemize}

These works use tasks, single-turn answers, full runs, aligned stages, or code edits as their paired or analytical units. Our unit of analysis is the frozen system. We pair comparisons on fixed states within multi-step games and trace how multiple components jointly affect the system's control flow.

\subsection{Failure Attribution}

\begin{itemize}
  \item \textbf{Who\&When}~\citep{zhang2025whowhen} defines a decisive error step counterfactually: replacing the step with a correct action changes the outcome from failure to success. In practice, however, both human annotations and three LLM-based judgments are made by inspecting logs. On manually constructed systems, the Hybrid method achieves 12.28\% step-level accuracy.
  \item \textbf{Agentic Attribution}~\citep{qian2026attribution} fixes a target action, progressively reveals historical prefixes at the component level, and measures changes in the action's likelihood. It also perturbs sentence-level context by deletion and retention to identify the components and sentences driving the action across eight trajectories.
\end{itemize}

We instead judge the task effect of each action with independent rules, and our intervention target is the frozen artifact set.

\subsection{Games and Rule-Based Environments}

\begin{itemize}
  \item \textbf{GAMEBoT}~\citep{lin2025gamebot} decomposes each turn in eight games into two or three subproblems and checks intermediate reasoning against ground truth generated by rule-based algorithms. In its experiments, GPT-4o achieves the highest average intermediate score, 0.52.
  \item \textbf{\citet{jung2026harness}} incrementally adds manually designed harness layers in noisy Battleship and measures win rate. Adding the second layer raises the win rate from 50.0\% to 74.1\%; this layer does not call an LLM.
\end{itemize}

We use rule-based verifiability in Gomoku to diagnose artifacts produced through natural evolution.

\subsection{Design-Element Comparison}\label{app:related-table}

In Table~\ref{tab:related-compare}, the ``same-state action pairing'' column indicates whether actions or outputs with and without an artifact are compared on the same decision state with a verified-identical input.

\begin{table*}[!t]
\centering
\footnotesize
\setlength{\tabcolsep}{4pt}
\renewcommand{\arraystretch}{1.15}
\caption{Design elements compared with the closest related work.}
\label{tab:related-compare}
\begin{tabularx}{\textwidth}{>{\raggedright\arraybackslash}p{2.5cm}>{\raggedright\arraybackslash}p{2.1cm}ZZZ>{\raggedright\arraybackslash}p{2.0cm}}
\toprule
Work & Artifact source & Pairing or intervention unit & Same-state action pairing & Primary readout & Cross-version \\
\midrule
EvoPathBench~\citep{lin2026evopath} & Evolutionary artifacts & Artifact on/off on held-out episodes & No (episode-level; trajectories diverge) & Episode risk-adjusted score difference & Yes (successive checkpoints) \\
Agent Plasticity~\citep{singh2026plasticity} & Evolutionary artifacts & Checkpoint-level games; LLM-observed annotations & Single case (appendix only) & Game score; usage annotations & Yes (successive checkpoints) \\
\citet{lin2026harnessbenefit} & Evolutionary artifacts & Agent $\times$ evolver cross-product & No & Pass-rate gain; loading and following rates & No \\
Misevolution~\citep{shao2026misevolve} & Hand-constructed (memory experiment) & Same query with and without memory & Yes & Safety label & Yes (before and after evolution) \\
SkillApt~\citep{guo2026skillapt} & Externally provided & Same task state with and without a skill & Yes (single answer) & Answer correctness; token cost & No \\
CMA~\citep{zhang2026cma} & Policy-internal history & Two histories crossed on a verified-identical input & Yes & Task branch selected by the action; pairwise reliability & No \\
\citet{huang2024selfcorrect} & No persistent artifact & Same problem before and after correction & Yes (single problem) & Incorrect-to-correct and correct-to-incorrect transitions & No \\
One Skill Too Many~\citep{yan2026oneskill} & Externally provided & Skill configurations for the same request & No & Skill use; core-function satisfaction rate & No \\
\citet{shang2026backfires} & Evolutionary artifacts & Single-skill A/B replay; lineage deletion & No & pass@1 & Yes (round by round) \\
VRL-Bench~\citep{bai2026vrlbench} & Evolutionary artifacts & Cases sharing the first round & No (diverging from round 2) & Case-level success rate & No \\
SAGE~\citep{wang2026sage} & Evolutionary artifacts & Incumbent and candidate on the same validation item & No (binary outcome comparison) & Per-item win/loss & Yes (adjacent versions) \\
ASSAY~\citep{wang2026assay} & Distilled external pipeline & Random masking & No & Mean difference in task success & No \\
\bottomrule
\end{tabularx}
\end{table*}

\end{document}